\documentclass{article}
\usepackage{PRIMEarxiv}
\pdfoutput=1
\usepackage[utf8]{inputenc} % allow utf-8 input
\usepackage[T1]{fontenc}    % use 8-bit T1 fonts
\usepackage{hyperref}       % hyperlinks
\usepackage{url}            % simple URL typesetting
\usepackage{booktabs}       % professional-quality tables
\usepackage{amsfonts}       % blackboard math symbols
\usepackage{nicefrac}       % compact symbols for 1/2, etc.
\usepackage{microtype}      % microtypography
\usepackage{lipsum}
\usepackage{fancyhdr}       % header
\usepackage{graphicx}       % graphics
\usepackage[nocompress]{cite}

\graphicspath{{media/}}     % organize your images and other figures under media/ folder

\usepackage[american]{babel}
\usepackage{mathtools} % amsmath with fixes and additions
\usepackage{tikz} % nice language for creating drawings and diagrams
\usepackage{times}
\usepackage{soul}
\usepackage{amsmath}
\usepackage{array}
\usepackage{amsthm}
\usepackage{algorithm}
\usepackage{algorithmic}
\usepackage[switch]{lineno}
\usepackage{subcaption}
\usepackage{multirow}
\usepackage{svg}
\usepackage{amssymb}  % 或者
\usepackage{amsfonts}
\usepackage{caption}
\usepackage{float}
\usepackage{indentfirst}
\usepackage{mathtools}
\usepackage{algorithm, algorithmic}
\title{TSCL:Multi-party loss Balancing scheme for deep learning Image steganography based on Curriculum learning
}
\author {
	FengChun Liu$^{1}$, Tong Zhang$^{2}$, Chunying Zhang$^{3}$\\
	$^{1}$Qianan College, North China University of Science and Technology, Tangshan, Hebei 063210, China\\
	$^{2}$School of Cyberspace Security, Beijing University of Posts and Telecommunications, Beijing 100876, China\\
	$^{3}$College of Science, North China University of Science and Technology, Tangshan, Hebei 063210, China\\
	\texttt{lnobliu@ncst.edu.cn, zenozt@bupt.edu.cn, hblg\_zcy@126.com}
}

\begin{document}
\maketitle

\begin{abstract}
For deep learning-based image steganography frameworks, in order to ensure the invisibility and recoverability of the information embedding, the loss function usually contains several losses such as embedding loss, recovery loss and steganalysis loss. In previous research works, fixed loss weights are usually chosen for training optimization, and this setting is not linked to the importance of the steganography task itself and the training process. In this paper, we propose a Two-stage Curriculum Learning loss scheduler (TSCL) for balancing multinomial losses in deep learning image steganography algorithms. TSCL consists of two phases: a priori curriculum control and loss dynamics control. The first phase firstly focuses the model on learning the information embedding of the original image by controlling the loss weights in the multi-party adversarial training; secondly, it makes the model shift its learning focus to improving the decoding accuracy; and finally, it makes the model learn to generate a steganographic image that is resistant to steganalysis. In the second stage, the learning speed of each training task is evaluated by calculating the loss drop of the before and after iteration rounds to balance the learning of each task. Experimental results on three large public datasets, ALASKA2, VOC2012 and ImageNet, show that the proposed TSCL strategy improves the quality of steganography, decoding accuracy and security.

\end{abstract}

% keywords can be removedin

% \keywords{First keyword \and Second keyword \and More}

\section{Introduction}\label{sec:intro}
Image steganography is a technique that protects the secure transmission of secret information by hiding information in the transmission media that is not recognizable by human vision, and plays an important role in practical communication, military and industrial scenarios. With the development of information security technology, steganography has gained a lot of research results, but with the development of steganalysis technology, steganography, in addition to the need to ensure the invisibility of information hiding, need to consider to improve the ability of steganographic images to avoid steganalysis detection. The invisibility of information embedding is guaranteed, i.e., the cost of information embedding on the original image is minimized. Existing adaptive steganography algorithms are designed based on the minimum embedding distortion framework, which employs coding to transform steganography into a problem of finding a better distortion function by constructing a distortion function. Such as using generative adversarial networks to automatically learn the embedding distortion cost and find the embedding location with the minimum distortion cost.Tang et al. proposed ASDL-GAN \cite{tang2017automatic}, which generates the embedding alteration probability map from the generator, puts the probability map into the embedding simulator to simulate the secret data embedding, generates the alteration location mapping map, and generates the secret-containing image by performing a point-and-point summation of the original image and the alteration location mapping map. Ensuring the security of information embedding, i.e., enhancing the ability of steganographic images to evade steganalysis. Researchers usually introduce adversarial training or adversarial attacks to enhance the security of information embedding, Yang et al \cite{yang2023acgis} proposed twin networks that preserve the noise residual relation in image sub-regions to generate adversarial samples to enhance the security of steganographic images.

With the rapid development of artificial intelligence technology, especially the wide application of deep learning in computer vision, natural language processing and other fields, it provides new ideas for steganography. In recent years, convolutional neural networks have been introduced to image steganography, transforming the traditional steganographic methods such as hand-designed features and a priori knowledge design into the steganographic method in which the convolutional neural network learns information embedding on its own. Image steganography models based on deep learning have been emerging, and some researchers have used generative adversarial networks to generate images suitable for steganography, or backpropagation of various deep learning models to automatically learn the embedding distortion cost, etc. Jing et al \cite{jing2021hinet} proposed the use of reversible neural networks for image steganography tasks, modeling image recovery explicitly as an inverse process of image hiding, which requires only one training of the network to obtain all the parameters of the hiding and recovering network, and achieves state-of-the-art performances in image recovery, and hiding invisibility. Subsequently, Xu et al \cite{xu2022robust} proposed a flow-based reversible neural network for the steganography task, which is easier to compute and adds structures such as content-aware noise projection to improve the robustness of steganographic images.

Deep Learning Image Steganography In order to ensure the invisibility of information embedding, security, and information recovery accuracy, the loss function will usually contain embedding loss, information recovery loss, and steganalysis loss. In previous studies, fixed loss weights are usually chosen for training optimization, and this setting is not linked to the actual training state of the model. Deep learning models in the actual training process, the learning state of the model is dynamically changing, and the fixed loss weights may not be able to meet the needs of adapting to the dynamic changes of the model. In addition, for the image steganography, the encoding loss is to learn the embedding of secret information, the decoding loss is to learn to reconstruct the secret information from the steganographic image, and the steganalysis loss is to learn the security of the steganographic image. Obviously, the first requirement satisfied by image steganography is to ensure the invisibility of the steganographic image, followed by information recovery accuracy and security.

In this work, we attempt to incorporate Curriculum learning ideas for optimizing multiple loss scenarios. The model is first made to focus on learning lossless message embedding, and then the training focus is shifted to message recovery accuracy and improving security. Curriculum learning is a machine learning training strategy that follows the introduction of different knowledge and concepts in different training phases, corresponding to a gradual increase in difficulty, so as to progressively master the learned knowledge. In this paper, a Two-stage Curriculum Learning loss scheduler (TSCL) is proposed. TSCL consists of two phases: a priori curriculum control and loss dynamics control. In the first stage, the model first focuses on learning the information embedding of the original image by controlling the loss weights in the multi-party adversarial training; secondly, the model shifts its learning focus to improving the decoding accuracy; and finally, the model learns to generate steganalysis-resistant steganographic images. In the second stage, the learning speed of each training task is evaluated by calculating the loss drop of the previous and subsequent iteration rounds to balance the learning of each task. The main contributions of this paper include the following two parts:

\begin{itemize}
\item[1] A multi-party loss-balanced scheduling method for deep Curriculum learning image steganography models is designed, including two phases of curriculum control and loss control.
\item[2] Experiments are conducted on three large public datasets, ALASKA2, VOC2012, and ImageNet, which show that the proposed TSCL method improves the quality of steganography, decoding accuracy, and security in steganography multi-loss scenarios.
\end{itemize}

\section{Related Work}

In this section, we first present research work on deep learning image steganography and multiple loss balancing. In addition, we address Curriculum learning algorithm concepts and related work.

\subsection{Deep learning steganography}

With the development of deep learning technology in recent years, neural networks produce many research results in fields such as computer vision, and researchers have gradually tried to use deep learning in the field of image steganography. Firstly, SGAN was proposed by Volkhonskiy et al \cite{volkhonskiy2016generative}, which takes random noise as input and generates carrier images suitable for steganography by DCGAN. Subsequently, Yu et al \cite{yu2021improved}, Li et al \cite{li2022gan} and Yuan et al \cite{yuan2022gan}improved the generative network, discriminative network and extraction network model structure of generative adversarial networks by introducing attention, adding cross-feedback channels and countering attacks to improve the information recovery accuracy and security. Su et al \cite{su2024stegastylegan} proposed a generalized image steganography framework, StyleGAN, designed to satisfy the goals of image security, capacity, and robustness through Distribution-Preserving Secret Data Modulator (DP-SDM) and Secret Data Extractor. However, this type of generative adversarial network based image steganography approach focuses more on image security and robustness with less steganographic capacity.

Embedded carrier-based image steganography schemes represented by various types of deep learning network models accomplish the embedding and extraction of secret information by utilizing neural networks on natural carrier images, which usually include information embedding networks, information recovery networks, steganalysis networks, and Discriminator. The SteGAN \cite{hayes2017generating} steganography model proposed by Hayes et al. defines a three-way adversarial game of encoding, decoding and steganalysis, which opens up a new research direction in the field of steganography, and its steganographic image can deceive the steganalysis network. Zhang et al \cite{zhang2019steganogan} defined a tripartite adversarial steganography framework for information hiding, information extraction and steganalysis in their study, which is widely used and studied due to its ability to achieve high capacity and enhance resistance to deep learning steganalysis. Subsequently, Peng et al \cite{peng2024image} and Yao et al \cite{yao2024high} introduced multi-scale convolutional module and multi-scale channel attention module under this steganography framework to improve the decoding accuracy, so that the model pays more attention to the features that are useful for improving the information embedding.

For larger capacity steganography research, Jing et al \cite{jing2021hinet} proposed the use of Invertible neural networks for image steganography tasks, modeling image restoration explicitly as an inverse process of image hiding, which requires only one training of the network to obtain all the parameters of the hiding and restoring network, and achieves the state-of-the-art performance in image restoration, hiding invisibility. This research has led to the widespread interest in Invertible neural networks for image steganography. Xu et al \cite{xu2022robust}, Feng et al \cite{feng2022image} and Li et al \cite{li2023iscmis} optimized Invertible neural networks by flow structure, swin transformer module, and spatial channel attention mechanism to enhance the invisibility and security of steganography and easier computation to guide the embedding of secret information into more secure image regions.

For loss function related research, Zhang \cite{zhang2019invisible} et al. proposed to hide the image in the Y-channel and proposed a loss function that combines structural similarity and multi-scale structural similarity to optimize the quality of steganography images. Subsequently, Feng et al \cite{jing2021hinet} and Li et al \cite{chekatamala2022analysis} proposed new low-frequency wavelet loss, FSIM loss which is closer to the image HVS perceptual model for improving image imperceptibility and security.

\subsection{Loss balancing}

When multiple losses are involved in the image steganography training process, it is important to consider how to trade-off multiple losses in the training process. In multi-task learning, which refers to learning multiple tasks in a single model to improve efficiency and generalization ability by sharing model parameters, this part of the research also involves the study of balancing multiple losses. Kendall et al \cite{kendall2018multi} proposed to utilize the uncertainty of the task as the loss weights, advocating that for tasks with higher losses, corresponding to higher uncertainty, the parameters should be updated with smaller gradients; on the contrary, for tasks with smaller losses, corresponding to lower uncertainty, the parameters should be updated with larger gradients, which avoids iterating the model in large step lengths to the wrong direction. Guo et al \cite{guo2018dynamic} proposed using the difficulty of the task as the basis for balancing the loss weights, claiming that the imbalance in difficulty between tasks can cause the model to focus on simple tasks and slow down or hinder the learning of difficult tasks. The study mentions dynamically ranking tasks in terms of difficulty during training and giving higher loss weights to difficult tasks.

From the degree of loss variation, Chen et al \cite{chen2018gradnorm} advocated that each task should be trained at a similar speed, and the gradient magnitude of different tasks should be utilized to correct the loss weights dynamically. For a task with a large magnitude of loss or when the training speed is too fast, its weights should be reduced to balance the learning of different tasks. Similar studies include Dynamic Weight Average (DWA) \cite{liu2019end}, which dynamically balances the loss weights based on the loss changes, takes into account the proportion of loss changes for each task, and is used as a measure of the difficulty of learning the task, and only calculates the loss of the task without accessing the gradient of the network, thus greatly saving memory overhead.

\subsection{Curriculum Learning}

Curriculum learning is a machine learning training strategy that follows the introduction of different knowledge and concepts at different training stages, corresponding to a gradual increase in difficulty, and thus a gradual mastery of the knowledge learned. The original Curriculum learning approach is a data-level training optimization strategy that guides the model to start learning from easier samples by categorizing the samples in terms of difficulty. As the model's ability improves, it gradually expands to more difficult and complex samples for training, allowing the model to spend less time on noisy and difficult samples in the pre-training phase, and guiding the model's training towards better local optima to achieve better generalization. Gong et al \cite{gong2023debiased} proposed a Debiased Contrastive Curriculum Learning (DCCL) strategy in the Person Re-Identification task. By improving the model's generalization ability for different domains through an easy-to-difficult training sequence, it makes the learning of unknown domains easier and effectively eliminates the domain bias problem. Farshbafan et al \cite{farshbafan2023curriculum} used the curriculum learning algorithm for optimizing Rain Streaks Removal, which guides the model to gradually learn the directionality of rain streaks, the appearance of rain streaks, and the image layers in a coarse-to-fine and easy-to-difficult manner.

With the development of research, researchers have given a broader definition of Curriculum learning in the process of application so that it can be applied to a wider and wider range of target tasks and domains. Existing Curriculum learning algorithms can be categorized into data level, model level and task level based on the target of application. Data-level Curriculum learning algorithms advocate model training starting with simple samples and gradually expanding to difficult samples as the model's capabilities increase. Kang et al. \cite{liao2023recrecnet} proposed Curriculum learning training based on Degree of Freedom (DoF) in a wide-angle image correction, training from similar transformation (4-DoF) to single-response transformation (8-DoF), which enables the network to learn more detailed deformations and accelerates model convergence.

Curriculum learning at the task level also follows the same idea of gradual progression by focusing on the relationships between tasks, approaching tasks incrementally, and using previously learned knowledge while training for the current task. Pentina \cite{pentina2015curriculum} et al. proposed to base the order of training on the relevance of the tasks, starting with tasks with high relevance, and transferring knowledge from previously learned tasks to the next task, rather than solving all tasks at the same time. Wang et al \cite{wang2020curriculum} proposed setting up curriculum tasks of increasing difficulty for training an end-to-end speech translation model, containing a curriculum pre-training method for transcription, comprehension and mapping, which gives the encoder the ability to generate the necessary features for the decoder. Model-based Curriculum learning allows the network model to achieve superior performance by regularly adapting the network model during training, such as setting up models with increasingly complex structures. Karras et al \cite{karras2017progressive} used to start with a low-resolution image from which the model captures the contour information of the data, and gradually add new network layers that process higher resolution details for increasing the detail information of the image in the subsequent training process.

\section{TSCL}

\subsection{Preliminaries}

The deep learning image steganography framework in this paper includes an encoding network $E(\cdot)$, a decoding network $D(\cdot)$, and a steganalysis network $S(\cdot)$. The encoding network performs information hiding, the decoding network performs information recovery, and the steganalysis network is used to evaluate the performance of the encoding network. The encoding network $E(\cdot)$ takes the original image and the secret message as input. Assuming that the inputs are the original image $C \in \mathbb{R}^{3\times H \times W}$ and the secret information $M \in \mathbb{R}^{D\times H \times W}$, the original image and the secret information are spliced over the channel as inputs to the coding network $I \in \mathbb{R}^{(3 +D)\times H \times W}$. After $E(\cdot)$ processing, the steganography image $C' \in \mathbb{R}^{3\times H \times W}$ is generated, and the process can be described as:
\begin{equation}
C'= E(C, M)
\end{equation}
The decoding network $D(\cdot)$ accepts the steganography image as input and generates the recovered secret message $M' \in \mathbb{R}^{D\times H \times W}$ from the steganography image, a process that can be described as:
\begin{equation}
M'= D(C')
\end{equation}
The main goal of the encoding network is to create a steganography image that keeps the visual appearance of the original image intact while preserving the secret information in the steganography image from being discovered. The steganalysis network $S(\cdot)$ takes the steganographic image as inputs and outputs a steganalysis score $s \in [0,1]$, which, as it gets closer to 1, indicates that there is a higher likelihood that the image contains the secret information:
\begin{equation}
s = S(C')
\end{equation}
In order to ensure the invisibility, recovery accuracy and security of information embedding, the loss function needs to contain multiple losses to optimize the training and improve the performance of information embedding in all aspects. The complete loss function contains information embedding loss, information recovery loss and steganalysis loss. The information embedding loss $L_{Encode}$ calculates the gap between the crypto-loaded image and the original image:
\begin{equation}
L_{E(\cdot)}(\theta_{E(\cdot)}, C) = d(C, C')
\end{equation}
$d(C, C')$ refers to the calculation of the distance between the original image and the carrier image, which is usually chosen to evaluate the similarity metrics of the two images. Information recovery loss calculates the gap between the original secret information and the recovered information:
\begin{equation}
L_{D(\cdot)}(\theta_{D(\cdot)}, M) = d(M, M') 
\end{equation}
Steganalysis loss is analyzed by calculating the likelihood that the secret information contained in a steganographic image will be discovered:
\begin{equation}
L_{S(\cdot)}(\theta_{S(\cdot)}, C, C') = d(s, label) 
\end{equation}
The $label$ refers to the label corresponding to each input image, which is 0 for the carrier image and 1 for the steganographic image, and is used to evaluate whether the steganographic image generated by the coding network can be easily detected by the steganalysis network. The complete loss function is as follows:
\begin{equation}
L_{total} = L_{E(\cdot)} + L_{D(\cdot)} + L_{S(\cdot)} 
\end{equation}

In previous research work, fixed loss weights are usually chosen to be directly summed for training optimization, and this setting is not linked to the actual training state of the model and the precedence of the steganography task. In addition to this, the learning state of the model is dynamically changing during the actual training of the deep learning model, and the fixed loss weights may not be able to meet the demand of adapting to the dynamic changes of the model.

In addition to this, for the specificity of the image steganography task, the encoding loss is for learning the embedding of secret information, the decoding loss is for learning to reconstruct the secret information from the steganographic image, and the steganalysis loss is for learning to improve the security of the steganographic image. The three losses correspond to the three tasks of learning information embedding, recovering secret information and resisting steganalysis. For image steganography, different tasks have different importance, we prefer the model to prioritize to perform better on certain important tasks, and the loss balancing method makes the model prioritize to focus on learning tasks with higher importance.

In this aspect, we try to introduce Curriculum learning for multinomial loss optimization in image steganography scenarios. In the early stage of training, we choose the idea of Curriculum learning for loss balancing, by controlling the three loss weights, so that the model firstly learns the information embedding in the first stage of curriculum control, then learns to reconstruct the complete secret information from the secret-loaded image, and finally, the model learns to resist the deep learning steganalysis. Second, different tasks have different learning difficulties. Different tasks with different optimization difficulties can lead to different convergence speeds for different tasks. Treating tasks equally may lead to learning overfitting for simple tasks, while tasks that are difficult to optimize may be underfitting during the learning process.

As a result, we perform loss weight balancing based on the change of each loss in the late stage of training. As a result, we propose a two-stage deep learning image steganography model multiparty loss balancing method based on curriculum learning, called Two-stage Curriculum Learning loss scheduler (TSCL). TSCL consists of two phases: a priori curriculum control and loss dynamics control. In the first stage, the model is firstly focused on learning information embedding by controlling the loss weights in multi-party adversarial training; secondly, the model is shifted to improve the decoding accuracy; and finally, the model is made to learn to generate steganalysis-resistant steganographic images. In the second stage, the learning rate of each training task is evaluated by calculating the loss drop of the previous and subsequent iteration rounds to balance the learning of each task. In the next two sections, we describe the curriculum control stage and the loss control stage

\subsection{curriculum control stage}

For the image steganography, the evaluation metrics can include imperceptibility, robustness, steganographic capacity, security, etc. Here the three losses of the multi-party adversarial steganography model correspond to the imperceptibility, decoding accuracy and security metrics. For steganography, imperceptibility should belong to the first to satisfy the indicators, if the classified image can not meet the imperceptibility, the classified image artifacts or distortion, that is, in the process of dissemination will be easy to be found, and can not achieve the purpose of the security of the transmission of secret information.

The second is that the receiver needs to reconstruct the secret information from the received encrypted image from the channel, and if the decoding accuracy is too low, the receiver cannot get the complete secret information from it. The last is the security metric, which measures whether the model-generated laden image can evade the detection of deep learning steganalysis. Therefore, a curriculum control method is chosen for managing the three loss weights so that the model focuses on learning a certain task during different training periods to satisfy a certain steganography metric first. This lossy curriculum scheduler is inspired by the idea of Curriculum learning and references the process of human learning by introducing different probabilities at different stages, which in turn leads to a gradual mastery of what has been learned. 

The training is divided into three steps, and the three loss weights are controlled to grow dynamically during the training process either by a function or by a fixed iteration step size. First, the weight of the coding loss is adjusted to the maximum threshold, at which time the gradient of this loss in the backpropagation update becomes correspondingly large, thus accounting for the largest proportion of parameter updates. At this point, the model focuses on learning the embedding of information in the early stage of training to ensure the imperceptibility of the encrypted image; then the weight of the decoding loss is adjusted to the maximum threshold, so that the model focuses on learning the reconstruction of information; finally, the weight of the steganalysis loss is adjusted to the maximum threshold, so that the model focuses on learning to enhance the security of the encrypted image.

\begin{figure}[h]
	\centering
	\centering
	\includegraphics[width=0.9\linewidth]{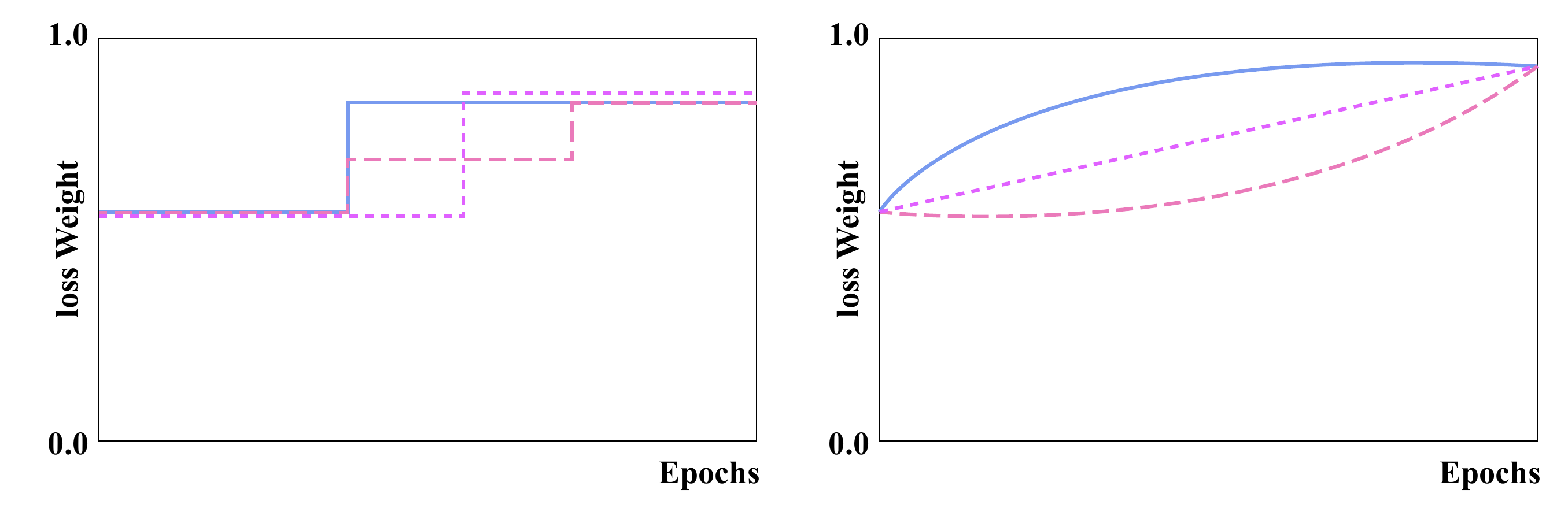}
	\caption{Discrete loss scheduling and continuous loss scheduling.}
	\label{fig1}
\end{figure}

Similar to the idea of applying Curriculum learning at the task level, learning all three losses at the same time or placing all three losses at the same importance for learning may result in slow convergence or poor performance of the model when the model is initially weakly fitted. In the control phase of this curriculum, the model focuses on learning a different task in each training period, whereas when entering the next step of the learning process, the model has already roughly learned the knowledge of the task corresponding to that loss from the previous step of learning, and still maintains the structure of the knowledge learned in the previous phase, i.e., keeping the weight of the corresponding loss from the previous phase does not decrease. In order to fully utilize this property, a variety of functions and iterative steps of the curriculum control scheme are selected, including two types of continuous loss scheduling and discrete loss scheduling, as shown in Figure. \ref{fig1}.

(1) Discrete Loss Scheduling. Discrete loss scheduling refers to loss weight adjustment at a fixed iteration step, where three losses are given fixed loss weights a priori at the beginning of training, and the other loss weights are adjusted after a fixed number of iterations with fixed values of the number of iterations. Four types of discrete loss scheduling strategies are designed, namely fixed iteration step size and fixed adjustment amplitude, fixed iteration step size and adjustment amplitude, and both fixed iteration step size and adjustment amplitude and unfixed iteration step size and adjustment amplitude, as shown in Figure. \ref{fig1}(a). As indicated by the pink dashed line, the iterative step size and weight ratio are fixed, the loss weight ratio is set to $K_1$ at the beginning of the training period, and after $C_1$ rounds of training, the loss weight ratio is adjusted to $K_1 + \alpha$; after $2\times C_1$ rounds of training, the loss weight ratio is adjusted to $K_1 + 2\alpha$, and so on.

(2) Continuous loss scheduling. Continuous-type loss scheduling refers to the continuous adjustment of the weight proportion of the loss during the training process, utilizing a function for the control of the weight coefficients to make a reasonable transition. The continuous loss scheduling in this part can be regarded as a function $f(t)$ mapping the number of iteration periods $t$ to a scalar $\lambda\in[0,1]$, which needs to satisfy monotonicity and incremental. The section was chosen to include convex functions with fast-to-slow growth rates, concave functions with slow-to-fast growth rates, linear functions, $a_0$ refers to the initial weight of a given loss, $a_1$ refers to the degree to which a given loss is controlled by the prior curriculum function, $C_1C_2C_3$ refers to the number of training iterations, $\alpha$ refers to the parameter of the weight of a given loss for a given training period, and $a_2$ refers to the parameter of the weight of a given loss for a given period of training up to the point of final training to converged loss weight parameter.
where the growth rate is a sinusoidal function from fast to slow, which can be described as equation (8):
\begin{equation}
\alpha = 
\begin{cases}
	a_0, epoch < C_1 \\
	a_0 + a_1 \times\sin(\min(\frac{\pi}{2}, \frac{epoch-C_1}{C_2-C_1} \times \frac{\pi}{2})), C_1 \leq epoch < C_2  \\
	a_2, epoch \geq C_2
\end{cases}
\end{equation}

linear function, which can be described as equation (9):
\begin{equation}
\alpha = 
\begin{cases}
	a_0, epoch < C_1 \\
	a_0 + a_1 \times(\frac{epoch-C_1}{C_2-C_1}), C_1 \leq epoch < C_2  \\
	a_2, epoch \geq C_2
\end{cases}
\end{equation}
An exponential function of the rate of growth from slow to fast, which can be described as equation (10):
\begin{equation}
\alpha = 
\begin{cases}
	a_0, epoch < C_1 \\
	a_0 + a_1 \times(e^{\frac{epoch-C_1}{C_2-C_1}} -1), C_1 \leq epoch < C_2  \\
	a_2, epoch \geq C_2
\end{cases}
\end{equation}

\subsection{loss control stage}

Different tasks are differently difficult to learn and different tasks are differently difficult to optimize, which can lead to different tasks converging at different rates. Treating tasks equally may lead to overfitting of simple tasks for learning, while tasks that are difficult to optimize may be underfitted during the learning process. In the first stage, the model optimizes each of the three losses with different weights, and this part of the training adjustment is due to a fixed setting, while the actual model training process is dynamic and may not be able to meet the needs of adapting to the dynamic changes of the model. As a result, after completing the first stage of curriculum control, we added a stage of weight adjustment based on the loss decline ratio. In this stage, we chose to adjust the training loss weights based on the loss change ratio of each round, and the process can be described as equation (11):
\begin{equation}
W_{k=1,2,3}(t) = \frac{Loss(t-1)}{Loss(t-2)}
\end{equation}
The loss decline rate $W_{k=1,2,3}(t)$, which refers to the ratio of the loss of the current $t-1$ round to the $t-2$ round, is used to measure the learning difficulty of the training. When this ratio is larger, it indicates that the training of that task is currently slow in decreasing the loss, suggesting that it is difficult to optimize and the next round of training should be given a larger loss weight. In addition to this, the weight product coefficient of each loss is introduced for a priori balanced initialization, as shown in equation (12):
\begin{equation}
D_{Encode} > D_{Decode} > D_{steganaylsis}
\end{equation}
This a priori coefficient is used to balance the dominance of the different tasks, and in the second stage, where the main task is still to ensure the imperceptibility of the carrier-tight image, multiplying the rate of loss decline in each round by the a priori coefficient can be described as equation (13):
\begin{equation}
\lambda_k(t) = D_k \times W_k(t)_{k=1,2,3}
\end{equation}
During training, when the three losses fall at approximately the same rate, i.e., when the three losses are optimized at approximately the same rate, the model is still more focused on training the coded losses to ensure imperceptibility due to the presence of a priori coefficients.

\section{Experiments}

\subsection{Experimental Platform and Datasets}

All the experiments in this paper are implemented in Linux system environment using Pytorch 1.10.1 deep learning framework, and the GPU is NVIDIA GeForce RTX 3090. The datasets used are three large public datasets, ImageNet, ALASKA2 and ImageNet. ImageNet is a large public computer vision dataset from which 25k images were extracted, of which 20k were used as the training set and the rest were used for testing. alaska2 is the public dataset for the ALASKA2 Image ALASKA2 is a public dataset for the ALASKA2 Image Steganalysis competition on the Kaggle platform, from which 10k original images are selected as the training set, 3k as the validation set, and 7k as the test set in the ALASKA2 dataset “Cover”.VOC2012 is a dataset used for target detection and semantic segmentation. 13k images to form the training set and the remaining 5k as the test and validation sets. Due to computational arithmetic limitations, all the original images of the dataset are processed to 128×128 pixels by Matlab program.

\subsection{Parameters}
Experiments were conducted using Adaptive Moment Estimation (Adam) to optimize the encoding and decoding networks with an initial learning rate of 0.001 and betas set to (0.9, 0.999). Stochastic Gradient Descent (SGD) algorithm was chosen to optimize the steganalysis adversarial network with an initial learning rate (lr) of 0.0001/3, weight\_decay set to 1e-8, and the parameters of the steganalysis network were updated every 5 batch. Each training Batchsize is 8, and the maximum number of iterations of the model is 120. The base model is an encoder-decoder-steganalysis network based on convolutional modules. The encoding network contains 9 layers of convolutional modules, and the decoding network contains 5 layers of convolutional modules, each of which contains a $3 \times 3$ convolution, BN, LeakyReLU. The steganalysis network was adopted from Zhang et al\cite{zhang2019invisible}.

The multinomial loss function set contains encoding loss, decoding loss and steganalysis loss. The Structural Similarity (SSIM), multi-scale structural similarity index (MSSSIM) and root mean square error (RMSE) are chosen as evaluation indexes for encoding loss, and the corresponding scale factors are $0.5,0.5,0.3$. Binary cross entropy was chosen for decoding loss and steganalysis loss. The steganography capacity in the experiment is D=1-3bpp (i.e., a $128\times128$ image hides information with a tensor of $128\times128\times$D). The loss control prior coefficients $D_{Encode}$,$D_{Decode}$,$D_{steganaylsis}$ are set to 1, 0.8, and 0.4.

\subsection{Experimental results}
1) TSCL scheme validation. The model trained without the fixed weight coefficients of TSCL is selected as the baseline scheme, and the performance of the model is compared with that of the model using the TSCL scheme, and the test results at 1-3 bpp hidden capacity are shown in Table \ref{tab1}. As shown in Table \ref{tab1}, the TSCL optimized model is on par with the baseline scheme in terms of imperceptibility, and the PSNR indexes are higher than those of the baseline scheme, and the TSCL optimized model has a certain degree of improvement in decoding accuracy, which is close to 100\% in 1-2 bpp steganography capacity, and the decoding accuracy is effectively improved with the increase of steganography capacity. The “steganalysis” score is the steganalysis test using a trained steganalysis model. The output range of the steganalysis is [0,1], and the closer the value is to 1, the greater the possibility that the image contains secret information, and the lower the security. As can be seen from Table 1, the security of the steganography images generated by the model trained by the TSCL scheme is improved on the ALASKA2 dataset and slightly decreased on the ImageNet dataset.
	
\begin{table}[htbp]
	\centering
	\caption{Comparison experiments between TSCL and baseline schemes}
	\label{tab1}
	\renewcommand{\arraystretch}{1.3}
	\begin{tabular}{ccccccccc}
		\hline
		Dataset & D & Scheme & SSIM & MSSSIM & PSNR & RMSE & Accuracy & Steganalysis \\
		\hline
		\multirow{6}{*}{ALASKA2} 
		& \multirow{2}{*}{1} & Baseline & 0.98351 & 0.99771 & 33.788 & 0.020 & 0.99 & 0.455 \\
		&  & TSCL & \textbf{0.99625} & \textbf{0.99901} & \textbf{37.227} & \textbf{0.014} & \textbf{1} & 0.342 \\
		\cline{2-9}
		& \multirow{2}{*}{2} & Baseline &\textbf{0.99547}  & \textbf{0.99909} & 35.086 & 0.017 & 0.99 & 0.402 \\
		&  & TSCL & 0.99434 & 0.99894 & \textbf{37.186} & \textbf{0.014} & \textbf{1} & 0.442 \\
		\cline{2-9}
		& \multirow{2}{*}{3} & Baseline & \textbf{0.99489} & \textbf{0.99877} & \textbf{34.857} & \textbf{0.018} & 0.82 & 0.441 \\
		&  & TSCL & 0.99176 & 0.99708 & 34.106 & 0.020 & \textbf{0.92} & 0.422 \\
		\hline
		\multirow{6}{*}{VOC2012} 
		& \multirow{2}{*}{1} & Baseline & 0.97616 & 0.99716 & 32.134 & 0.024 & 0.99 & 0.434 \\
		&  & TSCL & \textbf{0.99225} & \textbf{0.99950} & \textbf{38.346} & \textbf{0.012} & 0.99 & 0.457 \\
		\cline{2-9}
		& \multirow{2}{*}{2} & Baseline & \textbf{0.99326} & \textbf{0.99922} & \textbf{35.976} & 0.016 & 0.99 & 0.458 \\
		&  & TSCL & 0.99030 & 0.99867 & 34.028 & 0.020 & 0.99 & 0.457 \\
		\cline{2-9}
		& \multirow{2}{*}{3} & Baseline & \textbf{0.99404} & 0.99908 & 36.546 & 0.015 & 0.92 & 0.357 \\
		&  & TSCL & 0.99350 & \textbf{0.99929} & \textbf{37.455} & \textbf{0.013} & 0.92 & 0.449 \\
		\hline
		\multirow{6}{*}{ImageNet} 
		& \multirow{2}{*}{1} & Baseline & 0.99017 & 0.99914 & 36.612 & 0.014 & 0.96 & 0.362 \\
		&  & TSCL & \textbf{0.99225} & \textbf{0.99950} & \textbf{38.346} & \textbf{0.012} & \textbf{1} & 0.414 \\
		\cline{2-9}
		& \multirow{2}{*}{2} & Baseline & \textbf{0.99348} & 0.99843 & 33.613 & 0.021 & 0.99 & 0.325 \\
		&  & TSCL & 0.99040 & \textbf{0.99876} & \textbf{35.582} & \textbf{0.016} & 0.99 & 0.391 \\
		\cline{2-9}
		& \multirow{2}{*}{3} & Baseline & 0.99091 & 0.99852 & 33.775 & 0.020 & 0.78 & 0.413 \\
		&  & TSCL & \textbf{0.99305} & \textbf{0.99917} & \textbf{35.003} & \textbf{0.017} & \textbf{0.88} & 0.423 \\
		\hline
	\end{tabular}
\end{table}

2) Comparison of curriculum control function schemes. In order to verify the effectiveness of the proposed continuous curriculum control strategy, the curriculum control phase strategy was used as a test alone.
The models containing sine, cosine, linear and exponential functions were chosen as tests, and the model without any curriculum control function for weight balancing was chosen as the baseline scheme. The test results for 3bpp steganography capacity on the dataset ALASKA are shown in Table \ref{tab2}.

\begin{table}[htbp]
	\centering
	\caption{Comparison experiment of curriculum control function scheme}
	\label{tab2}
	\renewcommand{\arraystretch}{1.3}
	\begin{tabular}{lcccc}
		\hline
		Function & SSIM & MSSSIM & PSNR & Accuracy \\
		\hline
		Baseline & \textbf{0.99489} & \textbf{0.99877} & 34.857 & 0.82 \\
		Sin function & 0.99434 & 0.99894 & \textbf{37.182} & 0.62 \\
		Cosine function & 0.98992 & 0.99714 & 34.119 & 0.83 \\
		Exponential function & 0.99463 & 0.99780 & 32.783 & \textbf{0.87} \\
		Linear function & 0.99311 & 0.99788 & 33.241 & 0.85 \\
		\hline
	\end{tabular}
\end{table}

As can be observed from Table \ref{tab2}, the sine function is a monotonic function of growth rate from fast to slow, the model quickly turns up the encoding loss weights in the early training period, the model controlled by the sine function has improved in steganography quality, the PSNR index is improved by 6\%, but the decoding accuracy has been decreased, which may be due to the failure of decoding loss to be efficiently learnt in the middle of the training period or when the loss is learnt slowly. The models controlled by cosine, exponential and linear functions were slightly lower in steganography quality than the baseline model, but the decoding accuracy was somewhat improved. It can be shown that using curriculum control alone and ignoring the progress and changes of the model during training can lead to degradation in the performance of some metrics.

3) Comparison of discrete curriculum control schemes. In order to verify the effectiveness of the proposed discrete curriculum control strategy, the curriculum control phase strategy was tested separately as a test. A scheme containing fixed iteration step size, fixed weight adjustment magnitude, fixed iteration step size and weight adjustment magnitude, fixed iteration step size and weight adjustment magnitude, and unfixed iteration step size and weight adjustment magnitude was selected as a test, and a model without any curriculum control function for weight balancing was selected as a baseline scheme. The results of the 3bpp steganography capacity test on the dataset ALASKA are shown in Table \ref{tab3}.

\begin{table}[htbp]
	\centering
	\caption{Comparative experiments of discrete schemes for curriculum control}
	\label{tab3}
	\renewcommand{\arraystretch}{1.3}
	\begin{tabular}{lcccc}
		\hline
		Discrete scheme & SSIM & MSSSIM & PSNR & Accuracy \\
		\hline
		Baseline & 0.99489 & \textbf{0.99877} & 34.857 & 0.82 \\
		Fixed the iteration step size & 0.99342 & 0.99836 & 35.171 & 0.64 \\
		Fixed adjustment weight amplitude & 0.98408 & 0.99667 & 33.017 & 0.56 \\
		The iteration step size and weight adjustment amplitude are fixed & \textbf{0.99548} & 0.99858 & \textbf{36.619} & 0.83 \\
		The iteration step size and weight adjustment amplitude are not fixed & 0.99298 & 0.99810 & 36.532 & \textbf{0.87} \\
		\hline
	\end{tabular}
\end{table}

As can be observed from Table \ref{tab3}, the model of the curriculum control scheme with both iterative step length and weight adjustment magnitude not fixed has the best performance overall, with several steganography image quality metrics higher than the baseline model, and the decoding accuracy is improved by 6\% compared to the baseline model. The curriculum control scheme with both fixed iteration step size and weight adjustment magnitude had slightly better steganography quality metrics than the baseline scheme, but lower decoding accuracy than the curriculum control scheme model with both fixed iteration step size and weight adjustment magnitude. The performance of the curriculum control scheme with fixed iteration step length and fixed adjustment of weight magnitude was degraded in both imperceptibility and decoding accuracy, suggesting that manually set loss weight adjustments are not applicable to deep Curriculum learning image steganography schemes. The training of neural network models is dynamic, and for different datasets and model structures, different task losses correspond to different training difficulties and progress, so it is not possible to determine the optimal regulation scheme manually a priori, so it is necessary to incorporate metrics that can indicate the progress of the training to be used in conjunction with the loss adjustment.

4) Comparison of the effectiveness of the two phases of TSCL. In order to verify the effectiveness of curriculum control and loss dynamic control in the two phases, the training programs with only curriculum control and only loss dynamic control are selected for testing and compared with the training programs without TSCL and with TSCL, respectively. The results of the 3bpp steganography capacity test on the dataset ALASKA2, Pascal VOC2012 are shown in Table \ref{tab4}.

\begin{table}[htbp]
	\centering
	\caption{Testing the effectiveness of curriculum control stage and loss dynamic control stage}
	\label{tab4}
	\renewcommand{\arraystretch}{1.3}
	\begin{tabular}{ccccccccc}
		\hline
		Dataset & \begin{tabular}[c]{@{}c@{}}Curriculum\\control\\stage\end{tabular} & \begin{tabular}[c]{@{}c@{}}Loss\\dynamic\\control\\stage\end{tabular} & SSIM & MSSSIM & PSNR & RMSE & Accuracy & Steganalysis \\
		\hline
		\multirow{4}{*}{ALASKA2} 
		& $\times$ & $\times$ & 0.99489 & 0.99877 & 34.857 & 0.0182 & 0.82 & 0.441 \\
		& $\times$ & $\sqrt{}$ & 0.99252 & 0.99775 & 34.803 & 0.0185 & 0.85 & 0.456 \\
		& $\sqrt{}$ & $\times$ & \textbf{0.99698} & \textbf{0.99946} & \textbf{38.278} & \textbf{0.0123} & 0.85 & 0.429 \\
		& $\sqrt{}$ & $\sqrt{}$ & 0.99350 & 0.99929 & 37.455 & 0.0134 & \textbf{0.92} & \textbf{0.422} \\
		\hline
		\multirow{4}{*}{VOC2012} 
		& $\times$ & $\times$ & 0.99404 & 0.99908 & 36.546 & 0.0151 & \textbf{0.92} & \textbf{0.357} \\
		& $\times$ & $\sqrt{}$ & \textbf{0.99551} & 0.99842 & 35.681 & 0.0167 & 0.86 & 0.419 \\
		& $\sqrt{}$ & $\times$ & 0.99190 & \textbf{0.99941} & \textbf{38.140} & \textbf{0.0124} & 0.90 & 0.418 \\
		& $\sqrt{}$ & $\sqrt{}$ & 0.99350 & \textbf{0.99929} & 37.455 & 0.0134 & \textbf{0.92} & 0.449 \\
		\hline
	\end{tabular}
\end{table}

As can be observed from Table \ref{tab4}, the model optimized by the TSCL scheme outperforms the baseline scheme model in several steganography quality metrics as well as decoding accuracy and security, among which the PSNR metrics are improved by 2.4\% to 7\% compared to the baseline model, which indicates that the proposed TSCL scheme can improve the performance of all aspects of the multi-party adversarial-based image steganography model by reasonably regulating the loss weights. The curriculum-only control scheme model outperformed the baseline scheme model for steganography quality assessment on both datasets, and the decoding accuracy metrics on the ALASKA2 dataset were slightly higher than those of the baseline scheme model, but still lower than those of the TSCL scheme model. The metrics for assessing steganography quality were lower than the baseline scheme model on both datasets for the loss-control-only scheme model, the decoding accuracy metrics were slightly higher than the baseline scheme model on the ALASKA2 dataset, and the model performance was lower than that of the TSCL scheme model in all aspects. The TSCL model is able to maintain some improvement in both steganography quality and decoding accuracy compared to the baseline scheme, whereas the scheme model with only curriculum control and loss control only ensures an improvement in a single metric, while the other metrics decrease. The experimental results demonstrate the effectiveness and practicality of the proposed two-stage control scheme for the TSCL scheme, which requires prioritizing the learning of encoding losses through the curriculum control a priori ordering model in the early stage to learn through a step-by-step approach, and then calculating the magnitude of the loss change in the later stage to measure the learning difficulty of the task corresponding to each loss.

\begin{figure}[htbp]
	\centering
	\centering
	\includegraphics[width=0.8\linewidth]{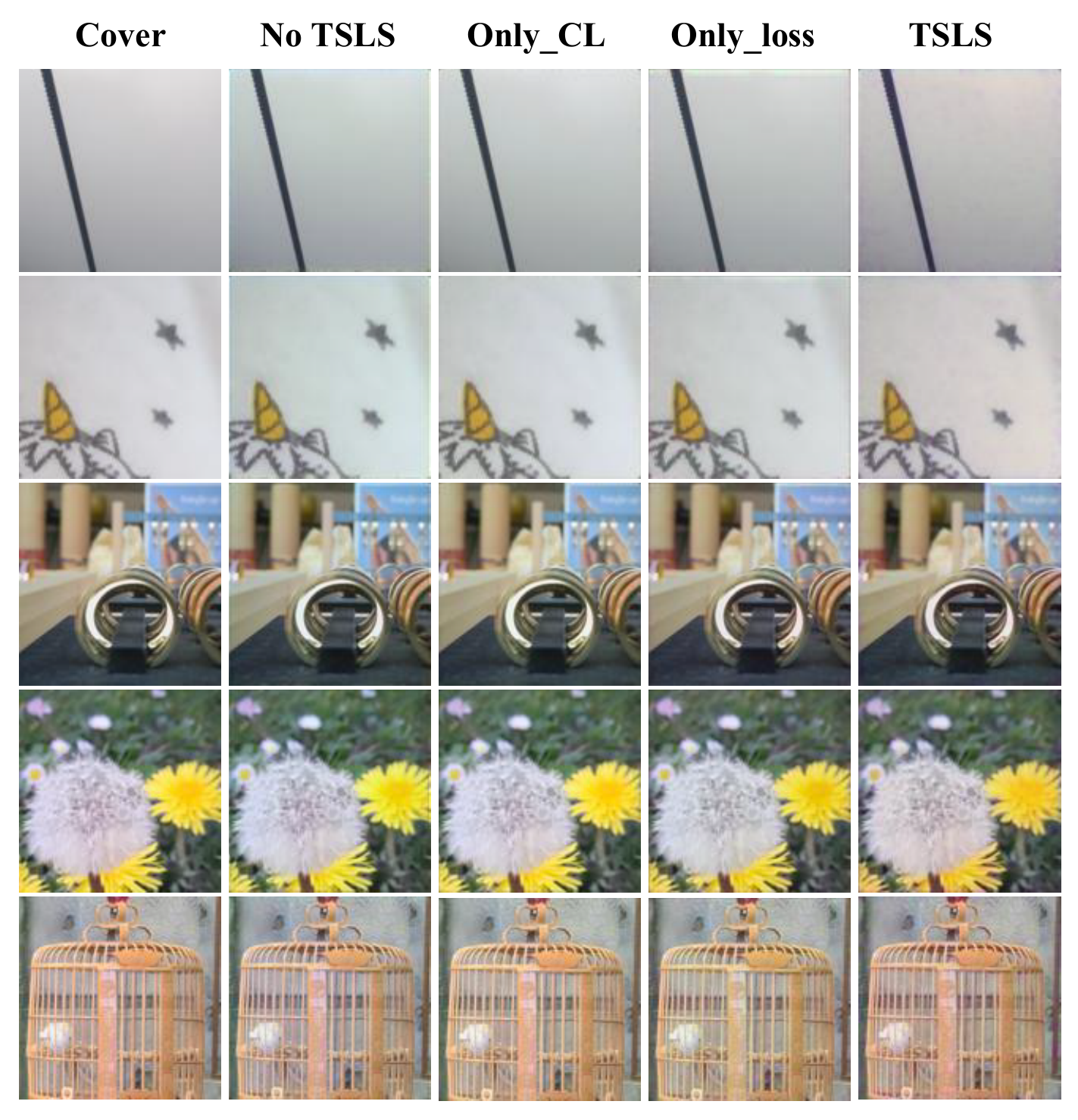}
	\caption{Comparison between TSCL scheme and steganography images using only curriculum control and only loss control model.}
	\label{fig2}
\end{figure}

5) Image visualization test. In order to verify the necessity of the two-stage training control in the TSCL training scheme to improve the quality of steganography images, the randomly trained baseline model “No TSCL”, the first-stage curriculum control only “Only CL”, the second-stage loss control only “Only Loss”, and the TSCL training scheme model “Only Loss” were chosen for comparison. Only CL”, ‘Only Loss’ with second stage loss control, and TSCL training scheme model for comparison. The test results are shown in Figure \ref{fig2}. As can be seen from Figure. \ref{fig2}, the randomly trained baseline model “No TSCL”, the first-stage curriculum control-only “Only CL”, the second-stage loss control-only “Only Loss” and the three comparison schemes are compared in the multi-stage loss control scenario. “The three comparison schemes all show the phenomena of greenish and yellowish on multiple test images, and there is a large difference in color with the original image. The steganography images generated by the TSCL training model are closer in structure and color, indicating the necessity of the two-stage training control in the TSCL training scheme to improve the quality of steganography.

\begin{figure}[htbp]
	\centering
	\centering
	\includegraphics[width=1\linewidth]{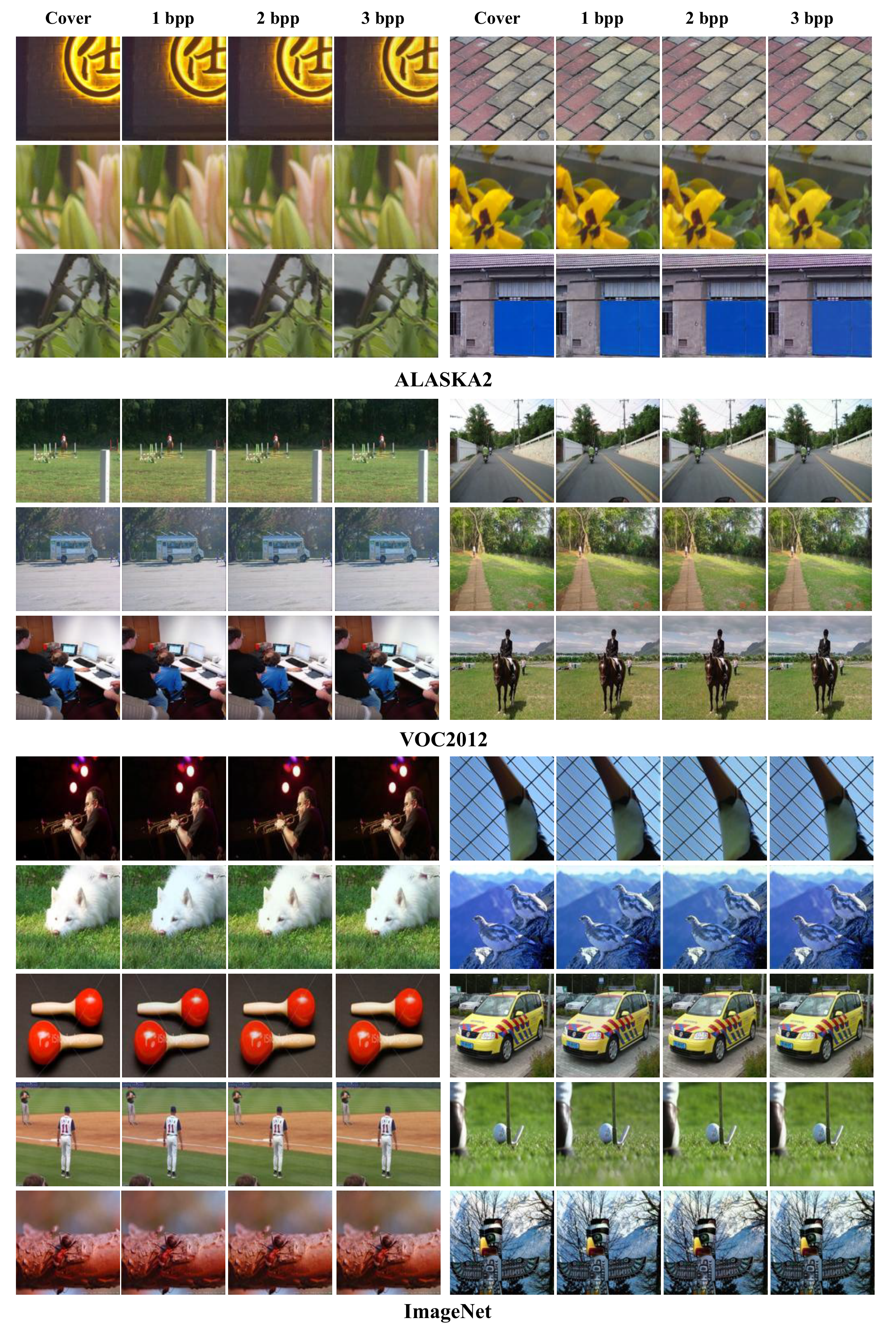}
	\caption{Comparison of cover image and stego image of TSCL scheme under 1-3 bpp capacity steganogayphy.}
	\label{fig3}
\end{figure}

In order to investigate the effect of the proposed TSCL on the model-generated loaded images, some images from the test set are selected for testing. The 1-3 bpp steganography capacity of the load-secret image and the original image are shown in Figure. \ref{fig3}. As can be seen from Figure. \ref{fig3}, the test images show that the loaded images and the original images are more similar in visual appearance and have very little difference in color and brightness, indicating that the TSCL scheme does not destroy the visual integrity of the images.

\begin{figure}[htbp]
	\centering
	\centering
	\includegraphics[width=1\linewidth]{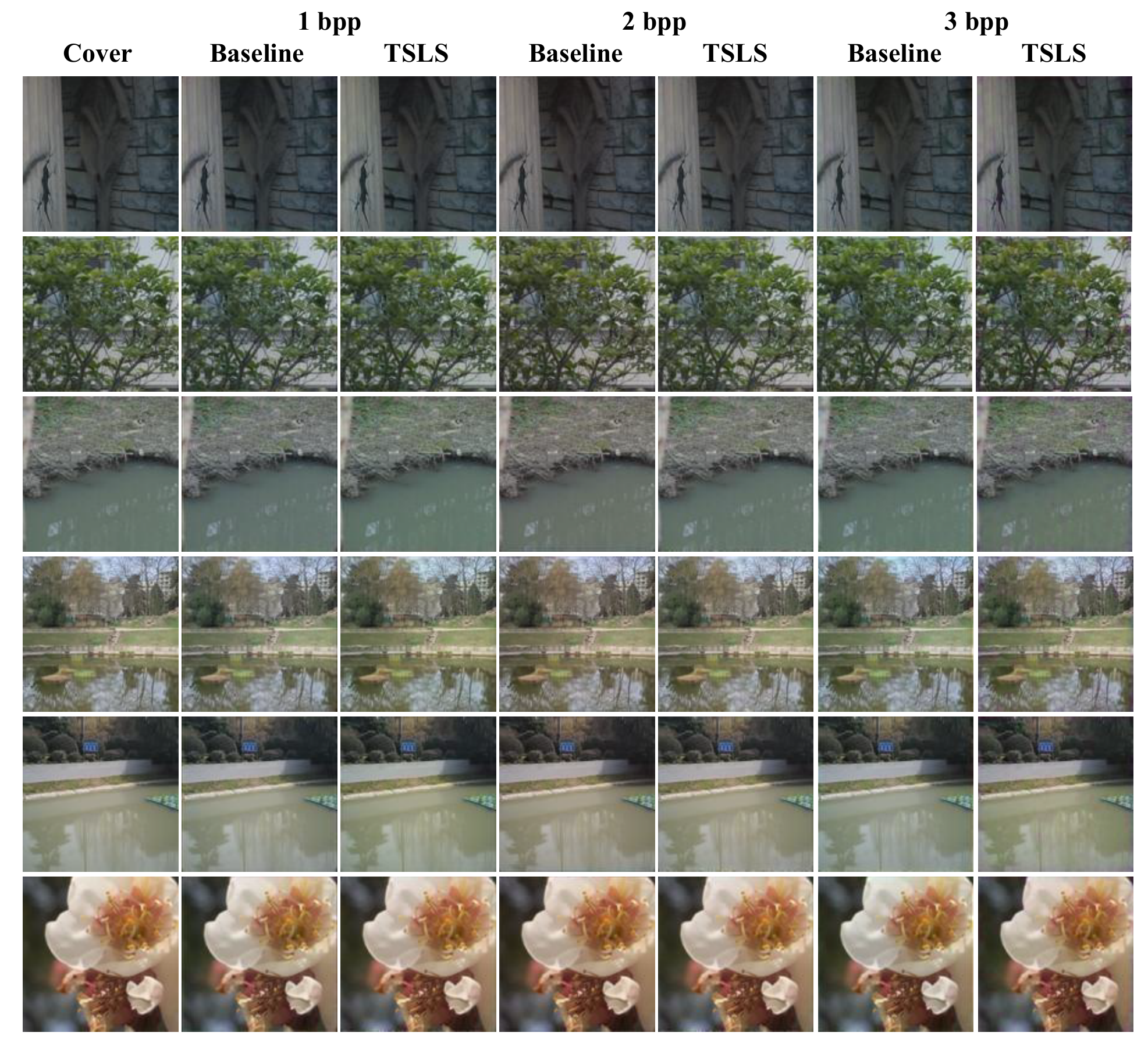}
	\caption{Comparison of image steganography quality between TSCL scheme and baseline scheme model.}
	\label{fig4}
\end{figure}

In order to further investigate the effectiveness of the TSCL model in improving the quality of image steganography, the randomly trained model of the baseline scheme and the TSCL-trained model are selected for comparison at 1-3 bpp steganography capacity. Multiple images are selected from the test set of three datasets, and steganography images are generated by the baseline model and the TSCL model, and the demonstrated results of the original images and steganography images are shown in Figure. \ref{fig4}. As can be seen from Figure. \ref{fig4}, the steganography images generated by the baseline model under 2-3 bpp steganography capacity appear yellowish, and there is a more obvious difference between the original image and the original image in terms of color, while the steganography images generated by the TSCL training model are closer to the original image.

\clearpage

\section{Conclusion}
In this paper, we attempt to introduce a scenario where curriculum learning is used for multinomial loss optimization. The first phase follows the idea of curriculum learning algorithm, so that the model focuses on learning information embedding in the training process firstly, and on learning information recovery and improving security secondly, so that the model focuses on optimizing a certain task in the same period. In the second phase, the product of the loss ratio and the prior coefficients of the previous and subsequent iteration rounds are calculated to adjust the loss weights in order to achieve the purpose of balancing each learning task. In this paper, several datasets are selected to evaluate the performance of the method on improving the invisibility of cryptography, information recovery accuracy and security, confirming the effectiveness of the method on several metrics.

%Bibliography

%\printbibliography
\bibliographystyle{unsrt}  % 或者使用其他样式如：plain, ieeetr, alpha等
\bibliography{ref}  % 不需要.bib扩展名

% \bibliographystyle{unsrt}  
% \bibliography{references}  

\end{document}